\documentclass[10pt,twocolumn,letterpaper]{article}

\usepackage{cvpr}  
\usepackage{times}  
\usepackage{helvet}  
\usepackage{courier}  
\usepackage{url}  
\usepackage{graphicx}  
\usepackage{amsmath}
\usepackage{amssymb}
\usepackage{pifont}
\usepackage{soul}
\usepackage{bbm}
\usepackage{enumitem}
\usepackage{tabularx,tabulary}
\usepackage{algorithm2e}
\usepackage{algpseudocode}
\usepackage{amsmath,amssymb} 
\usepackage[table,dvipsnames,svgnames,x11names]{xcolor}
\usepackage{booktabs}
\usepackage{pdfpages}

\usepackage[pagebackref=true,breaklinks=true,colorlinks,bookmarks=false]{hyperref}

\usepackage{multirow}
\usepackage{makecell}
\usepackage{amsmath}
\usepackage{bm}

\definecolor{dgreen}{rgb}{0.04,0.7,0.13}
\newcommand{\cmark}{{\color{red}\ding{51}}}%
\newcommand{\xmark}{{\color{dgreen}\ding{55}}}%
\cvprfinalcopy 

\def\cvprPaperID{6639} 

\ifcvprfinal\fi

\begin{document}

\title{Self-Supervised 3D Human Pose Estimation via \\ 
Part Guided Novel Image Synthesis
}

\author{Jogendra Nath Kundu$^1$\thanks{Equal contribution.} \qquad Siddharth Seth$^1$\footnotemark[1] \qquad Varun Jampani$^2$ \qquad Mugalodi Rakesh$^1$ \\R. Venkatesh Babu$^1$ \qquad Anirban Chakraborty$^1$\\
$^1$Indian Institute of Science, Bangalore  \qquad $^2$Google Research\\
}

\maketitle
\thispagestyle{empty}

\begin{abstract}
Camera captured human pose is an outcome of several sources of variation. Performance of supervised 3D pose estimation approaches comes at the cost of dispensing with variations, such as shape and appearance, that may be useful for solving other related tasks. As a result, the learned model not only inculcates task-bias but also dataset-bias because of its strong reliance on the annotated samples, which also holds true for weakly-supervised models. Acknowledging this, we propose a self-supervised learning framework\footnote{Project page: {\url{http://val.cds.iisc.ac.in/pgp-human/}}} to disentangle such variations from unlabeled video frames. We leverage the prior knowledge on human skeleton and poses in the form of a single part based 2D puppet model, human pose articulation constraints, and a set of unpaired 3D poses. Our differentiable formalization, bridging the representation gap between the 3D pose and spatial part maps, not only facilitates discovery of interpretable pose disentanglement, but also allows us to operate on videos with diverse camera movements. Qualitative results on unseen in-the-wild datasets establish our superior generalization across multiple tasks beyond the primary tasks of 3D pose estimation and part segmentation. Furthermore, we demonstrate state-of-the-art weakly-supervised 3D pose estimation performance on both Human3.6M and MPI-INF-3DHP datasets. 

\end{abstract}


\begin{figure}[!tbhp]
\begin{center}
	\includegraphics[width=1.00\linewidth]{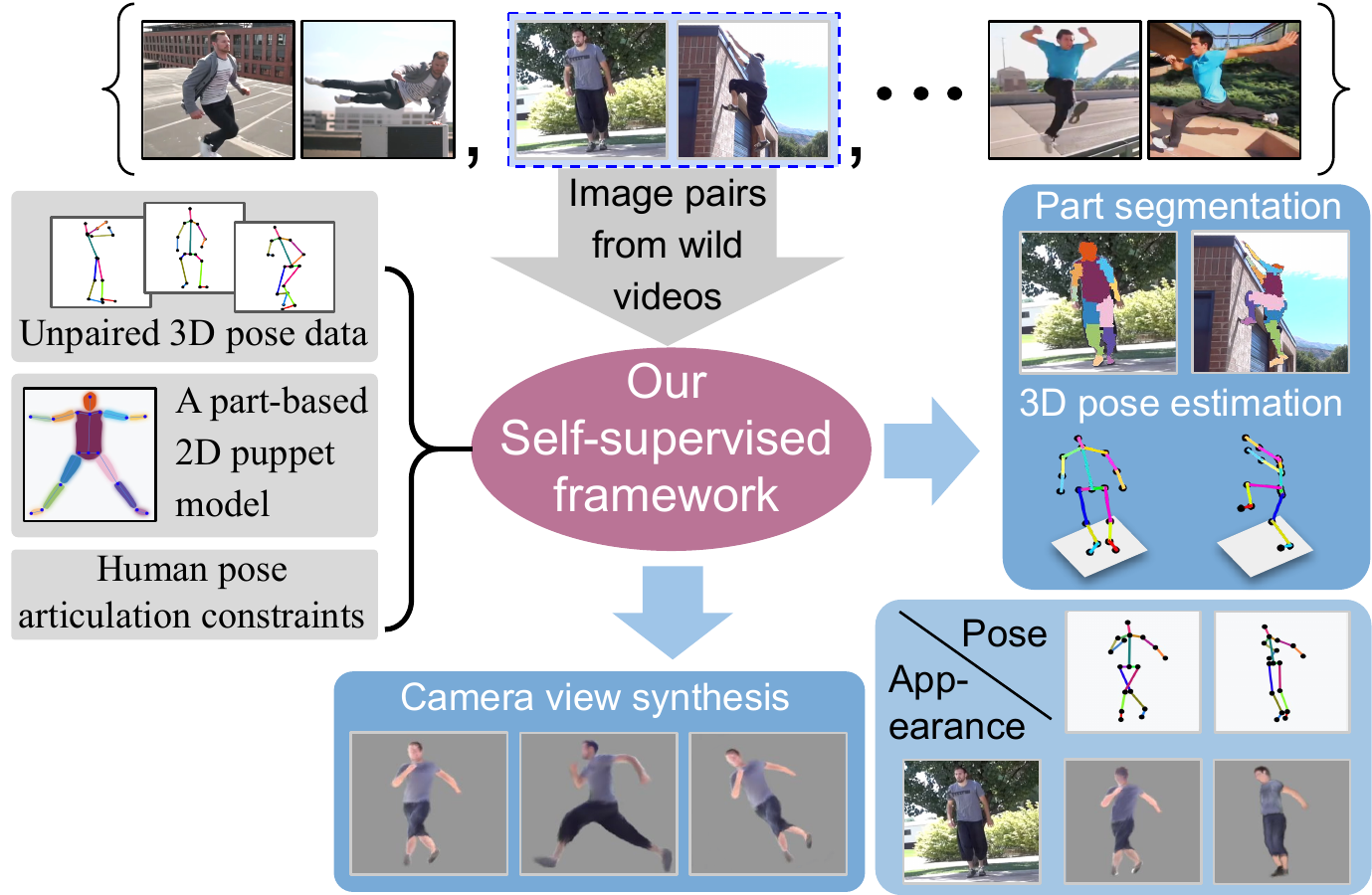}
	\vspace{-3.8mm}
	\caption{ 
	Our self-supervised framework 
	not only produces 3D pose and part segmentation but also enables novel image synthesis via interpretable latent manipulation of the disentangled factors. 
	}
    \vspace{-6mm}
    \label{fig:concept}  
\end{center}
\end{figure}

\section{Introduction}
Analyzing humans takes a central role in computer vision systems.~Automatic estimation of 3D pose and 2D part-arrangements of 
highly deformable humans from monocular RGB images remains an important, challenging and unsolved problem. 
This ill-posed classical inverse problem has diverse
applications in human-robot interaction~\cite{Svenstrup:2009:PEA:1703775.1703969}, augmented reality~\cite{hagbi2010shape}, gaming industry, etc. 

In a fully-supervised setting~\cite{sun2018integral,moreno20173d,Dabral_2018_ECCV}, the advances in this area are mostly driven by 
recent deep learning architectures and the collection of large-scale annotated samples. However, unlike 2D landmark annotations, it is very difficult to manually annotate 3D human pose on 2D images.
A usual way of obtaining 3D ground-truth (GT) pose annotations is through a well-calibrated in-studio multi-camera setup~\cite{ionescu2013human3,sigal2010humaneva}, which is difficult to configure 
in outdoor environments. This results in a limited diversity in the available 3D pose datasets, which greatly limits the generalization of supervised 3D pose estimation models. 

To facilitate better generalization, several recent works~\cite{chen2019unsupervised,rhodin2018learning} leverage 
weakly-supervised learning techniques that reduce the need for 3D GT pose annotations. Most of these works use an auxiliary task such as multi-view 2D pose estimation to train a 3D pose estimator~\cite{chen2019weakly,kocabas2019self}. Instead of using 3D pose GT for supervision, a 3D pose network is supervised with loss functions on multi-view projected 2D poses. To this end, several of these works still require considerable annotations in terms of paired 2D pose GT~\cite{chen2019unsupervised,tung2017adversarial,novotny2019c3dpo,kong2019deep}, multi-view images~\cite{kocabas2019self} and known camera parameters~\cite{rhodin2018learning}. Dataset bias still remains a challenge in these techniques as they use paired image and 2D pose GT datasets which have limited diversity. Given the ever-changing human fashion and evolving culture, the visual appearance of humans keeps varying and we need to keep updating the 2D pose datasets accordingly.

In this work, we propose a differentiable and modular self-supervised learning framework for monocular 3D human pose estimation along with the discovery of 2D part segments. Specifically, our encoder network takes an image as input and outputs 3 disentangled representations: 1. view-invariant 3D human pose in canonical co-ordinate system, 2. camera parameters and 3. a latent code representing foreground (FG) human appearance. Then, a decoder network takes the above encoded representations, projects them onto 2D and synthesizes FG human image while also producing 2D part segmentation. Here, a major challenge is to disentangle the representations for 3D pose, camera, and appearance. We achieve this disentanglement by training on video frame pairs depicting the same person, but in varied poses. We self-supervise our network with consistency constraints across different network outputs and across image pairs. Compared to recent self-supervised approaches that either rely on videos with static background~\cite{rhodin2018unsupervised} or work with the assumption that temporally close frames have similar background~\cite{jakab2018unsupervised}, our framework is robust enough to learn from large-scale in-the-wild videos, even in the presence of camera movements. We also leverage the prior knowledge on human skeleton and poses in the form of a single part-based 2D puppet model, human pose articulation constraints, and a set of unpaired 3D poses. Fig.~\ref{fig:concept} illustrates the overview of our self-supervised learning framework.



{Self-supervised learning from in-the-wild videos is challenging due to diversity in human poses and backgrounds in a given pair of frames which may be further complicated due to missing body parts.} 
We achieve the ability to learn on these wild video frames
with a pose-anchored deformation of puppet model
that bridges the representation gap between the 3D pose and the 2D part maps in a fully differentiable manner. 
In addition, the part-conditioned appearance decoding allows us to reconstruct only the FG human appearance resulting in robustness to changing backgrounds.

Another distinguishing factor of our technique 
is the use of well-established pose prior constraints. In our self-supervised framework, we explicitly
model 3D rigid and non-rigid pose transformations by adopting a differentiable parent-relative local limb kinematic model, thereby reducing 
ambiguities in the learned representations. In addition, for the predicted poses to follow the real-world pose distribution, we make use of an unpaired 3D pose dataset. 
We interchangeably use predicted 3D pose representations and sampled real 3D poses during training to guide the model towards a plausible 3D pose distribution.

Our network also produces useful part segmentations. With the learned 3D pose and camera representations, we model depth-aware inter-part occlusions resulting in robust part segmentation. To further improve the segmentation beyond what is estimated with pose cues, we use a novel differentiable shape
uncertainty map that enables extraction of limb shapes from the FG appearance representation.

We make the following main contributions:
\begin{itemize}



\vspace{-2mm}
\item 
We present techniques to explicitly constrain the 3D pose by modeling it at its most fundamental form of rigid and non-rigid transformations. This
results in interpretable 3D pose predictions, even in the absence of any auxiliary 3D cues such as multi-view or depth.

\vspace{-2mm}
\item 
We propose a differentiable part-based representation which enables us to selectively attend to foreground human appearance which in-turn makes it possible to learn on in-the-wild videos with changing backgrounds in a
self-supervised manner.

\vspace{-2mm}
\item We demonstrate generalizability of our self-supervised framework on \textit{unseen} in-the-wild datasets, such as LSP~\cite{johnson2010clustered} and YouTube. Moreover, we achieve \textit{state-of-the-art} weakly-supervised 3D pose estimation performance on both Human3.6M~\cite{ionescu2013human3} and MPI-INF-3DHP~\cite{mehta2017monocular} datasets against the existing approaches.
\end{itemize}




\section{Related Works}
\label{sec:related-works}

Human 3D pose estimation is a well established problem in computer vision, specifically in fully supervised paradigm. Earlier approaches~\cite{ramakrishna2014pose,yang2011articulated,tompson2014joint,chen2014articulated} proposed to infer the underlying graphical model for articulated pose estimation. However, the recent CNN based approaches~\cite{bulat2016human,newell2016stacked,VNect_SIGGRAPH2017} focus on regressing spatial keypoint heat-maps, without explicitly accounting for 
the underlying limb connectivity information. However, the performance of such models heavily relies on a large set of paired 2D or 3D pose annotations. 
As a different approach, \cite{katircioglu2018learning} proposed to regress latent representation of a trained 3D pose autoencoder to indirectly endorse a plausibility bound on the output predictions. Recently, several weakly supervised approaches utilize varied set of auxiliary supervision other than the direct 3D pose supervision (see Table~\ref{tab:char}).
In this paper, we address a more challenging scenario where we consider access to only a set of unaligned 2D pose data to facilitate the learning of a plausible 2D pose prior (see Table~\ref{tab:char}).

\begin{table}[t]
	\footnotesize
	\caption{  
	Characteristic comparison of our approach against prior weakly-supervised human 3D pose estimation works, in terms of access to direct (paired) or indirect (unpaired) supervision levels. \vspace{0.8mm} }
	\centering
	\setlength\tabcolsep{3.0pt}
	\resizebox{0.47\textwidth}{!}{
	\begin{tabular}{l|ccc|c|c}
	\hline
 		\multirow{2}{*}{Methods} & \multicolumn{3}{c|}{ \makecell{Paired sup.\\(MV: muti-view)}} &
 		\multirow{2}{*}{\makecell{\vspace{-3.5mm}\\Unpaireed\\2D/3D pose\\Supervision}} &
 		\multirow{2}{*}{\makecell{\vspace{-5.5mm}\\Sup. for \\latent to\\3D pose\\ mapping}} \\
 		\cline{2-4}  
 		 & \makecell{MV\\ pair} & \makecell{Cam.\\ extrin.} & \makecell{2D \\pose} & 
 		 \\ \hline\hline
  		\rowcolor{gray!00}
 		Rhodin \etal~\cite{rhodin2018unsupervised} &\cmark &\cmark &\xmark &\xmark &\cmark 
 		\\
  		\rowcolor{gray!6}
 		Kocabas \etal~\cite{kocabas2019self} &\cmark &\xmark &\cmark &\xmark &\xmark     \\
  		\rowcolor{gray!00}
 		Chen \etal~\cite{chen2019weakly} &\cmark &\xmark &\cmark &\xmark &\cmark 
 		\\
  		\rowcolor{gray!6}
 		Wandt \etal~\cite{wandt2019repnet} &\xmark &\xmark &\cmark &\cmark &\xmark    \\
  		\rowcolor{gray!00}
 		Chen \etal~\cite{chen2019unsupervised} &\xmark &\xmark &\cmark &\cmark &\xmark    \\
  		\rowcolor{gray!14}
 		Ours &\xmark &\xmark &\xmark &\cmark &\xmark   \\ 
		\hline
	\end{tabular}}
	\vspace{-2mm}
	\label{tab:char}
\end{table} 

In literature, while 
several supervised shape and appearance disentangling techniques~\cite{balakrishnan2018synthesizing,ma2018disentangled,ma2017pose,esser2018variational,siarohin2018deformable} exist, the available unsupervised pose estimation works (\ie in the absence of multi-view or camera extrinsic supervision), are mostly limited to 2D landmark estimation~\cite{denton2017unsupervised,jakab2018unsupervised} for rigid or mildly deformable structures, such as  facial landmark detection, constrained torso pose recovery etc. The general idea~\cite{kanazawa2016warpnet,rocco2017convolutional,thewlis2017unsupervised,thewlis2017nips} is to utilize the relative transformation between a pair of images depicting a consistent appearance with varied pose. Such image pairs are usually sampled from a video satisfying appearance invariance~\cite{jakab2018unsupervised} or synthetically generated deformations~\cite{rocco2017convolutional}. 


Beyond landmarks, object parts~\cite{lorenz2019unsupervised} can infer shape alongside the pose. Part representations are best suited for 3D articulated objects as a result of its occlusion-aware property as opposed to simple landmarks. In general, the available unsupervised part learning techniques~\cite{singh2012unsupervised,hung2019scops} are mostly limited to segmentation based discriminative tasks. On the other hand, \cite{Yang2016end,novotny2017anchornet} explicitly leverage the consistency between geometry and the semantic part segments. However, the kinematic articulation constraints are well defined in 3D rather than in 2D~\cite{akhter2015pose}. 
Motivated by this, we aim to leverage the advantages of both non-spatial 3D pose~\cite{katircioglu2018learning} and spatial part-based representation~\cite{lorenz2019unsupervised} by proposing a novel 2D pose-anchored part deformation model. 

\begin{figure*}
\begin{center}
    \vspace{-2mm}
	\includegraphics[width=1.0\linewidth]{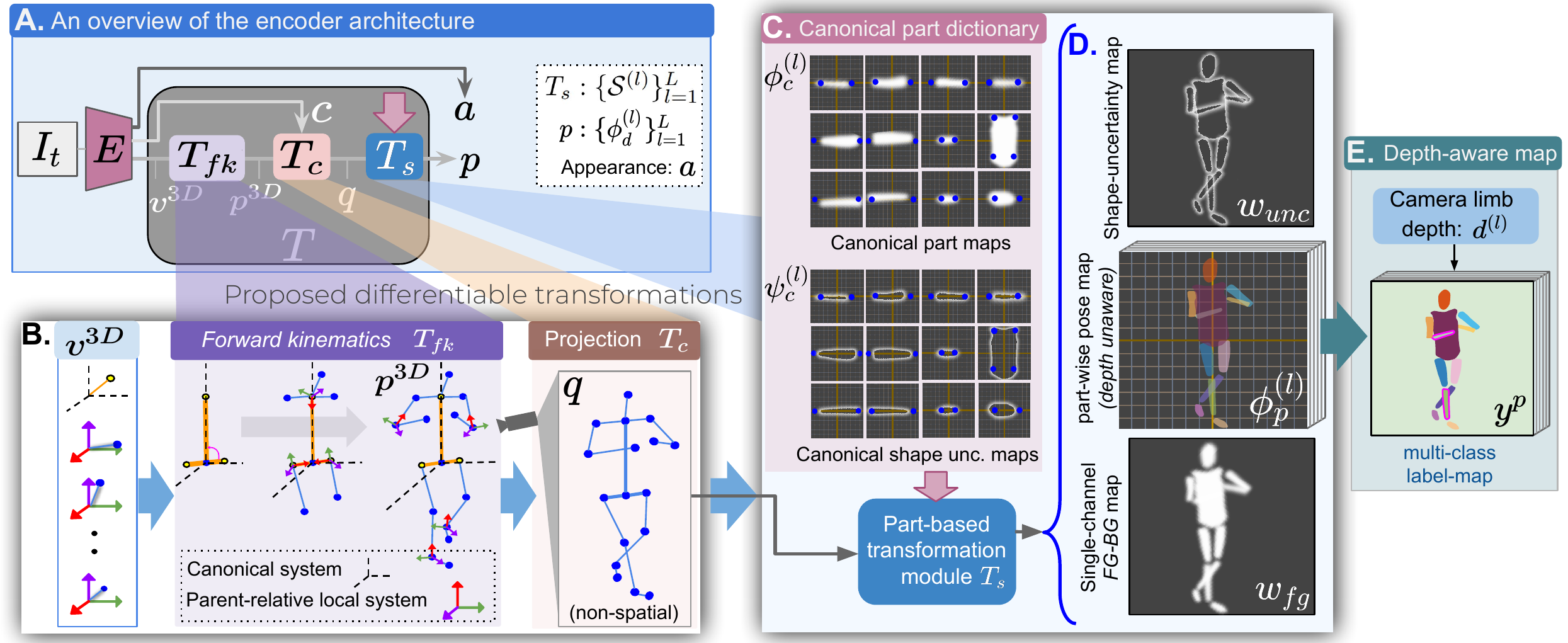}
	\caption{
	\textbf{A.} An overview of the full encoder module.
	\textbf{B.} Transforming the parent-relative local 3D vectors $v_{3D}$ to camera projected 2D pose $q$.
	\textbf{C.} The template 2D puppet model.
	\textbf{D.} Puppet imitates the pose in $q$.
	\textbf{E.} Image independent depth-aware part segmentation.
	}
 	\label{fig:main1}    
    \vspace{-6mm}
\end{center}
\end{figure*}



\section{Approach}
\label{sec:approach}

We develop a differentiable framework for self-supervised disentanglement of 3D pose and foreground appearance from in-the-wild video frames of human activity.
%


Our self-supervised 
framework builds on the conventional encoder-decoder architecture (Sec.~\ref{sec:3_2}). Here, the encoder produces a set of local 3D vectors from an input RGB image. This is then processed through a series of 3D transformations, adhering to the 3D pose articulation constraints to obtain a set of 2D coordinates (camera projected, non-spatial 2D pose). 
In Sec.~\ref{sec:3_1}, we define a set of part based representations followed by carefully designed differentiable transformations required to bridge the representation gap between the non-spatial 2D pose and the spatial part maps. 
This serves three important purposes. First, their spatial nature facilitates compatible input pose conditioning for the fully-convolutional decoder architecture. 
Second, it enables the decoder to selectively synthesize only FG human appearance ignoring the variations in the background. 
Third, it facilitates a novel way to encode the 2D joint and part association using a single template puppet model. 
Finally, Sec.~\ref{sec:3_3} describes the proposed self-supervised paradigm which makes use of the pose-aware spatial part maps for simultaneous discovery of 3D pose and part segmentation using image pairs from wild videos.

\subsection{Joint-anchored spatial part representation}\label{sec:3_1}
One of the major challenges in unsupervised pose or landmark detection is to map the model-discovered landmarks to the standard landmark conventions. 
This is essential to facilitate the 
subsequent task-specific pipelines, which expect the input pose to follow a certain convention. Prior works~\cite{rhodin2018unsupervised,kundu2020ksp} rely on paired supervision to learn this mapping. In absence of such supervision, we aim to encode this convention in a \textit{canonical part dictionary} where the association of 2D joints with respect to the body parts is extracted from a \textit{single} manually annotated puppet template (Fig.~\ref{fig:main1}{\color{red}C}, top panel). This can be interpreted as a 2D human puppet model, which can approximate any human pose deformation via independent spatial transformation of body parts while keeping intact the anchored joint associations.

\vspace{1.5mm}
\textbf{Canonical maps.} We extract \textit{canonical part maps}, $\{\phi_c^{(l)}\}_{l=1}^L$ (here, $l$: limb index and $L$: total number of limbs or parts), where we perform erosion followed by Gaussian blurring of binary part segments to account for the associated \textit{shape} uncertainty (\ie body shape or apparel shape variations). We represent $\phi_c^{(l)}:\mathcal{U}\rightarrow \mathbb[0,1]$, 
where $\mathcal{U}\in\mathbb{N}^2$ is the space of spatial indices.  
In addition, we also extract \textit{canonical shape uncertainty maps} $\{\psi_c^{(l)}\}_{l=1}^L$ to specifically highlight only the uncertain regions (Fig.~\ref{fig:main1}{\color{red}C}, bottom panel). 
The two anchored joint locations for each limb $l$ and its corresponding part map $\phi_c^{(l)}$ are denoted as $r_c^{l(j_1)}$, $r_c^{l(j_2)}$ $\in\mathcal{U}$, except for the \textit{torso} which is represented using 4 
joints. 

\vspace{1.5mm}
\textbf{Part deformation model.} For a given 2D pose $q\in\mathbb{R}^{2J}$ with $J$ being the total number of joints, \textit{part-wise pose maps} are obtained as independent spatial-transformations of the \textit{canonical part maps}, \ie $\phi_p^{(l)}=\mathcal{S}^{(l)}\circ\phi_c^{(l)}$. Here, $\mathcal{S}^{(l)}$ represents an affine transformation of the spatial indices $u\in\mathcal{U}$, whose rotation, scale, 
and translation parameters are obtained as a function of $(q^{l(j_1)}, q^{l(j_2)}, r_c^{l(j_1)}, r_c^{l(j_2)})$, where $q^{l(j_1)}$, $q^{l(j_2)}$ denote the joint locations associated with the limb $l$ in pose $q$. Similarly, we also compute the \textit{part-wise shape uncertainty maps} as $\psi_p^{(l)}=\mathcal{S}^{(l)}\circ\psi_c^{(l)}$. Note that, $\{\phi_p^{(l)}\}_{l=1}^L$ and $\{\psi_p^{(l)}\}_{l=1}^L$ are unaware of inter-part occlusion in the absence of limb-depth information. Following this, we obtain single-channel maps (see Fig.~\ref{fig:main1}{\color{red}D}), \ie 

\textbf{a)} \textit{shape uncertainty map} as $w_{unc} = \max_l\psi_p^{(l)}$, and 

\textbf{b)} \textit{single-channel FG-BG map} as $w_{fg} = \max_l\phi_p^{(l)}$. 

The above formalization bridges the representation gap between the raw joint locations, $q$ and the output spatial maps $\phi_p$, $w_{fg}$, and $w_{unc}$, thereby facilitating them to be used as differentiable spatial maps for the subsequent self-supervised learning. 




\vspace{1.5mm}
\textbf{Depth-aware part segmentation.} For 3D deformable objects, a reliable 2D part segmentation can be obtained with the help of 
following attributes, \ie a) 2D skeletal pose, b) part-shape information, and c) knowledge of inter-part occlusion. Here, the 2D skeletal pose and the knowledge of inter-part occlusion can be extracted by accessing camera transformation of the corresponding 3D pose representation. Let, the depth of the 2D joints in $q$ with respect to the camera be denoted as $q_d^{l(j_1)}$ and $q_d^{l(j_2)}$.  
We obtain a scalar depth value associated with each limb $l$ as, $d^{(l)}= ( q_d^{l(j_1)} + q_d^{l(j_2)} )/2 $.  
%
We use these depth values to alter the strength of depth-unaware \textit{part-wise pose maps}, $\phi_p^{(l)}$ at each spatial location, $u\in \mathcal{U}$ by modulating the strength of part map intensity as being inversely proportional to the depth values. 
This is realized in the following steps: 

\textbf{a)} 
$\phi_d^{(l)}(u) 
= \textit{softmax}_{l=1}^{L} (\phi_p^{(l)}(u)/{d}^{(l)})$, 


\textbf{b)} $\phi_d^{(L+1)}(u) = 1 - \max_{l=1}^L {\phi}_d^{(l)}(u) $, and 

\textbf{c)}
$ \bar{\phi}_d^{(l)}(u) 
= \textit{softmax}_{l=1}^{L+1} ({\phi}_d^{(l)}(u))$. 

Here, $(L+1)$ indicates the spatial-channel dedicated for the background. 
Additionally, a non-differentiable 2D part-segmentation map (see Fig.~\ref{fig:main1}{\color{red}E}) is obtained as,

$y^p(u, l) = \mathbbm{1}(l = \text{argmax}_{l=1}^{L+1}\bar{\phi}_d^{(l)}(u))$. 

\subsection{Self-supervised pose network}\label{sec:3_2}
The architecture for self-supervised pose and appearance disentanglement consists of a series of pre-defined differentiable transformations facilitating discovery of a constrained latent pose representation. 
As opposed to imposing learning based constraints~\cite{habibie2019wild}, 
we devise a way around where the 3D pose articulation constraints (\ie knowledge of joint connectivity and bone-length) are directly applied via structural means, 
implying guaranteed constraint imposition. 


\begin{figure*}
\begin{center}
    \vspace{-1mm}
	\includegraphics[width=1.0\linewidth]{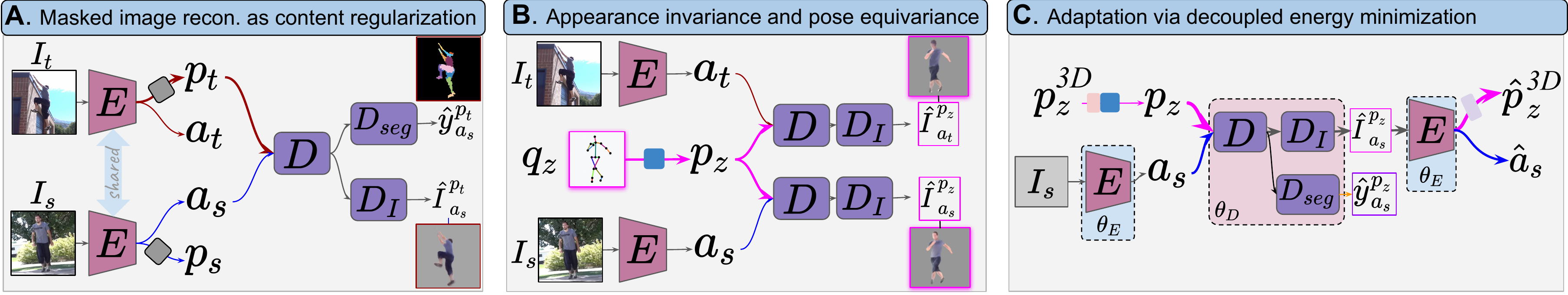}
	\vspace{-5.0mm}
	\caption{An overview of data-flow pipelines for the proposed self-supervised objectives. Images close to the output heads show the network output for the given set of exemplar input tuple. 
	Here, the color of transformation blocks are consistent with Fig.~\ref{fig:main1}{\color{red}A}. 
	}
 	\label{fig:main2}    
    \vspace{-6mm}
\end{center}
\end{figure*}

\noindent
\textbf{a) Encoder network.} As shown in Fig.~\ref{fig:main1}{\color{red}A}, the encoder $E$ takes an input image $I$ and outputs three disentangled factors, a) a set of local 3D vectors: $v^{3D}\in\mathbb{R}^{3J}$, b) camera parameters: $c$, and c) a FG appearance: $a\in\mathbb{R}^{H\times W\times \text{Ch}}$.

As compared to spatial 2D geometry~\cite{rhodin2018unsupervised,lorenz2019unsupervised}, discovering the inherent 3D human pose is a highly challenging task considering the extent of associated non-rigid deformation, and rigid camera variations~\cite{akhter2015pose,kundu2019unsupervised}. To this end, we define a canonical coordinate system $C$, where \textit{face-vector} of the skeleton is canonically aligned along the +ve X-axis, thus making it completely view-invariant. Here, the \textit{face-vector} is defined as the perpendicular direction of the plane spanning the neck, left-hip and right-hip joints. As shown in Fig.~\ref{fig:main1}{\color{red}B}, in $v^{3D}$, except the pelvis, neck, left-hip and right-hip, all other joints are defined at their respective parent relative local coordinate systems (\ie parent joint as the origin with axis directions obtained by performing Gram-Schmidt orthogonalization of the parent-limb vector and the \textit{face-vector}). Accordingly, we define a recursive forward kinematic transformation $T_{fk}$ to obtain the canonical 3D pose from the local limb vectors, \ie $p^{3D} = T_{fk}(v^{3D})$, which accesses a constant array of limb length magnitudes~\cite{zhou2016deep}. 

Here, the camera extrinsics, $c$ (3 rotation angles and 3 restricted translations ensuring that the camera-view captures all the skeleton joints in $p^{3D}$) is obtained at the encoder output, whereas a fixed perspective camera projection 
is applied to obtain the final 2D pose representation, \ie 
$q = T_c(p^{3D})$. A part based deformation operation on this 2D pose $q$ (Sec.~\ref{sec:3_1}) is shown as $p=T_s( q, d^{(l)})$, where $T_s:\{\mathcal{S}^{(l)}\}_{l=1}^{L}$ 
and $p: \{\phi_d^{(l)}\}_{l=1}^L$ (following the depth aware operations on $\phi_p^{(l)}$). Finally, $T$ denotes the entire series of differentiable transformations, \ie $T_{s}\circ T_c \circ T_{fk}$, as shown in Fig.~\ref{fig:main1}{\color{red}A}. Here, $\circ$ denotes composition operation.

\noindent
\textbf{b) Decoder network.} The decoder takes a concatenated representation of the FG appearance, $a$ and pose, $p$ as input to obtain two output maps, i) a reconstructed image $\hat{I}$, and ii) a predicted part segmentation map $\hat{y}$ via a bifurcated CNN decoder (see Fig.~\ref{fig:main2}{\color{red}A}). The common decoder branch, $D$ consists of a series of up-convolutional layers conditioned on 
the spatial pose map $p$ at each layer's input (\ie multi-scale pose conditioning).
Whereas, $D_I$ and $D_{seg}$ follow up-convolutional layers to their respective outputs. 

\subsection{Self-supervised training objectives}\label{sec:3_3}
The prime design principle of our self-supervised framework is to leverage the interplay between the pose and appearance information by forming paired input images of either consistent pose or appearance. 

Given a pair of source and target image, $(I_s, I_t)$, sampled from the same video \ie with consistent FG appearance, the shared encoder extracts their respective pose and appearance as $(p_s, p_t)$ and $(a_s, a_t)$ (see Fig.~\ref{fig:main2}{\color{red}A}). We denote the decoder outputs as $(\hat{I}_{a_s}^{p_t}, \hat{y}_{a_s}^{p_t})$ while the decoder takes in pose, $p_t$ with appearance, $a_s$ (this notation is consistent in later sections). Here, $\hat{I}_{a_s}^{p_t}$ is expected to depict the person in the target pose $p_t$. Such a cross-pose transfer setup is essential to restrict leakage of pose information 
through the appearance. Whereas, the low dimensional bottleneck of $v^{3D}$ followed by the series of differentiable transformations prevents leakage of appearance through pose~\cite{jakab2018unsupervised}. 

To effectively operate on wild video frames (\ie beyond the in-studio fixed camera setup~\cite{rhodin2018unsupervised, jakab2018unsupervised}), we aim to utilize the pose-aware, spatial part representations as a means to disentangle the FG from BG. 
Thus, we plan to reconstruct $\hat{I}_{a_s}^{p_t}$ with a constant BG color $\textit{BG}_c$ and segmented FG appearance (see Fig.~\ref{fig:main2}{\color{red}A}). Our idea stems from the concept of co-saliency detection~\cite{zhang2016co,hsu2018unsupervised}, where the prime goal is to discover the common or salient FG from a given set of two or more images. Here, the part appearances belonging to the model predicted 
part regions have to be consistent across $I_s$ and $I_t$ for a successful self-supervised pose discovery.


\textbf{Access to unpaired 3D/2D pose samples.} We denote $p^\textit{3D}_z$ and $q_z=T_c(p_z^\textit{3D})$ as a 3D pose and its projection (via random camera), sampled from an \textit{unpaired} 3D pose dataset $\mathcal{D}_z$, respectively. 
Such samples can be easily collected without worrying about BG or FG diversity in the corresponding camera feed (\ie a single person activity). We use these samples to further
constrain the latent space 
towards realizing a plausible 3D pose distribution.

\vspace{1mm}\noindent
\textbf{a) Image reconstruction objective.} 
Unlike~\cite{rhodin2018unsupervised,jakab2018unsupervised}, we do not have access to the corresponding ground-truth representation for the predicted $\hat{I}_{a_s}^{p_t}$, giving rise to an increased possibility of producing degenerate solutions or \textit{mode-collapse}. In such cases, the model focuses on fixed background regions as the common region between the two images, specifically for in-studio datasets with limited BG variations. One way to avoid such scenarios is to select image pairs with completely diverse BG (\ie select image pairs with high L2 distance in a sample video-clip). 

To explicitly restrict the model from inculcating such BG bias, we incorporate content-based regularization. An {uncertain} pseudo FG mask is used to establish a one-to-one correspondence between the reconstructed image $\hat{I}_{a_s}^{p_t}$ and the target image $I_t$. This is realized through a spatial-mask, $m_{sal}$ which highlights the salient regions (applicable for any image frame) or regions with diverse motion cues (applicable for frames captured in a fixed camera). We formulate a pseudo (\textit{uncertain}) reconstruction objective as, 

\vspace{-5mm}
$$
\mathcal{L}_I^{u} = (1-\hat{y}_{a_s}^{p_t}(L+1)+\beta m_{sal}^{I_t})    \otimes|\hat{I}_{a_s}^{p_t}-I_t|
$$

Here, $\otimes$ denotes pixel-wise weighing and $\beta$ is a balancing hyperparameter. Note that, 
the final loss is computed as average over all the spatial locations, $u\in\mathcal{U}$.
This loss enforces a self-supervised consistency between the pose-aware part maps, $\hat{y}_{a_s}^{p_t}$ and the salient common FG to facilitate a reliable pose estimate.

As a novel direction, we utilize $q_z\sim\mathcal{D}_z$ 
to form a pair of image predictions $(\hat{I}_{a_s}^{p_z}, \hat{I}_{a_t}^{p_z})$ following simultaneous appearance invariance and pose equivariance (see Fig.~\ref{fig:main2}{\color{red}B}). Here, a \textit{certain} reconstruction objective is defined as, 

\vspace{-5mm}
$$\mathcal{L}_I^{c} = w_{fg}^{p_z}\otimes|\hat{I}_{a_s}^{p_z} - \hat{I}_{a_t}^{p_z}| + (1-w_{fg}^{p_z})\otimes|\hat{I}_{a_s}^{p_z} - \textit{BG}_c|$$

\vspace{1mm}\noindent
\textbf{b) Part-segmentation objective.} 
Aiming to form a consistency between the true pose $p_z$ and the corresponding part segmentation output, we formulate, 

\vspace{-5mm}
$$
\mathcal{L}_{seg} = (1-w_{unc})\otimes\textit{CE}(\hat{y}_{a_s}^{p_z}, y^{p_z}) +w_{unc}\otimes\textit{SE}(\hat{y}_{a_s}^{p_z})$$


\begin{algorithm}[!b]
\vspace{-2mm}
\SetAlgoLined
$\theta_E$: Trainable parameters of the Encoder $E$\\   
$\theta_D$: Trainable parameters of the Decoder \\
$\:\:\:\:\:\:\;$ (includes $D$, $D_{I}$, and $D_{seg}$)

\For {$\textit{iter} < \textit{MaxIter}$}{
 \uIf{$\textit{iter}\pmod{2}\neq 0$}{
 Update $\theta_E$ by optimizing $\mathcal{L}_{p_z^\textit{3D}}$ and $\mathcal{L}_{a_s}$ in $\:$ separate \textit{Adagrad} optimizers on frozen $\theta_D$.
 }
 \Else{
 Update $\theta_D$ by optimizing $\mathcal{L}_{p_z^\textit{3D}}$ and $\mathcal{L}_{a_s}$ in $\:$ separate \textit{Adagrad} optimizers on frozen $\theta_E$.
 }
 
 
 Update $(\theta_E, \theta_D)$ by optimizing $\mathcal{L}_{I}^u$, $\mathcal{L}_{I}^c$, and $\:$ $\mathcal{L}_{seg}$ in separate \textit{Adagrad} optimizers.
 }
 \vspace{1mm}
\caption{Self-supervised learning with the proposed adaptation via decoupled energy minimization. 
\vspace{-2mm}} \label{algo:1}
\end{algorithm}

Here, $\textit{CE}$, and $\textit{SE}$ denote the pixel-wise cross-entropy and self-entropy respectively. Moreover, we confidently enforce segmentation loss with respect to one-hot map $y^{p_z}$ (Sec.~\ref{sec:3_1}) only at the \textit{certain} regions, while minimizing the Shanon's entropy for the regions associated with \textit{shape uncertainty} as captured in $w_{unc}$. Here, the limb depth required to compute $y^{p_z}$ is obtained from $\hat{p}_z^{\textit{3D}} = E_p(\hat{I}_{a_s}^{p_z})$ (Fig.~\ref{fig:main2}{\color{red}C}).

In summary, the above self-supervised objectives form a consistency among $p$, $\hat{y}$, and $\hat{I}$; 

\textbf{a)} $\mathcal{L}_I^u$ enforces consistency between $\hat{y}$ and $\hat{I}$, 

\textbf{b)} $\mathcal{L}_{I}^c$ enforces consistency between $p$ (via $w_{fg}$) and $\hat{I}$, 

\textbf{c)} $\mathcal{L}_{seg}$ enforces consistency between $p$ (via $y^{p_z}$) and $\hat{y}$. 

However, the model inculcates a discrepancy between the predicted pose and the true pose distributions. It is essential to bridge this discrepancy as $\mathcal{L}_I^c$ and $\mathcal{L}_{seg}$ rely on true pose $q_z=T_c(p_z^\textit{3D})$, whereas $\mathcal{L}_I^u$ relies on the predicted pose $q_t$. Thus, we employ an adaptation strategy to guide the model towards realizing a plausible pose prediction.


\vspace{1mm}\noindent
\textbf{c) Adaptation via energy minimization.} 
Instead of employing an ad-hoc adversarial discriminator~\cite{chen2019unsupervised,yang20183d}, we devise a simpler yet effective decoupled energy minimization strategy~\cite{han2018co,jaderberg2017decoupled}. 
We avoid a direct encoder-decoder interaction during gradient back-propagation, by updating
%
%
the encoder parameters, while freezing the 
decoder parameters and vice-versa. However, this is performed while enforcing a reconstruction loss at the output of the \textit{secondary} encoder in a cyclic auto-encoding scenario (see Fig.~\ref{fig:main2}{\color{red}C}). The two energy functions are formulated as $\mathcal{L}_{p^\textit{3D}_z} = |p^\textit{3D}_z-\hat{p}^\textit{3D}_z|$ and $\mathcal{L}_{a_s} = |a_s-\hat{a}_s|$, where $\hat{p}_z^{\textit{3D}} = 
E_p(\hat{I}_{a_s}^{p_z})$ and $\hat{a}_s = E_a(\hat{I}_{a_s}^{p_z})$.

The decoder parameters are updated to realize a faithful $\hat{I}_{a_s}^{p_z}$, as the frozen encoder expects $\hat{I}_{a_s}^{p_z}$ to match its input distribution of real images (\ie $I_s$) for an effective energy minimization. Here, the encoder can be perceived as a frozen energy network as used in energy-based GAN~\cite{zhao2016energy}. A similar analogy applies while updating the encoder parameters with gradients from the frozen decoder. Each alternate energy minimization step is preceded by an overall optimization of the above consistency objectives, where both encoder and decoder parameters are updated simultaneously (see Algo.~\ref{algo:1}).
 


\section{Experiments}

We perform a thorough experimental analysis to establish the effectiveness of our proposed framework on 3D pose estimation, part segmentation and novel image synthesis tasks, across several datasets beyond the in-studio setup.



\vspace{1mm}
\noindent
\textbf{Implementation details.} We employ an ImageNet trained \textit{Resnet-50} architecture~\cite{he2016deep} as the base CNN for the encoder $E$. We first bifurcate it into two CNN branches dedicated to pose and appearance, then the pose branch is further bifurcated into two multi-layer fully-connected networks to obtain the local pose vectors $v^{3D}$ and the camera parameters $c$. 
While training, we use separate AdaGrad optimizers~\cite{duchi2011adaptive} for each loss term at alternate training iterations. 
We perform appearance (color-jittering) and pose augmentations (mirror flip and inplane rotation) selectively for $I_t$ and $I_s$ conceding their invariance effect on $p_t$ and $a_s$ respectively.

\vspace{1mm}
\noindent
\textbf{Datasets.} We train the \textit{base-model} on image pairs sampled from a mixed set of video datasets \ie Human3.6M~\cite{ionescu2013human3} (H3.6M) and an in-house collection of in-the-wild YouTube videos (YTube). As opposed to the in-studio H3.6M images, the YTube dataset constitutes a substantial diversity in apparels, action categories (dance forms, parkour stunts, etc.), background variations, and camera movements. The raw video frames are pruned to form the suitable image pairs after passing them through an off-the-shelf person-detector~\cite{ren2015faster}. We utilize an unsupervised saliency detection method~\cite{zhu2014saliency} to obtain $m_{sal}$ for the wild YTube frames, whereas for samples from H3.6M $m_{sal}$ is obtained directly through the BG estimate~\cite{rhodin2018unsupervised}. Further, LSP~\cite{LassnerClosing} and MPI-INF-3DHP~\cite{mehta2017monocular} (3DHP) datasets are used to evaluate generalizability of our framework. 
{We collect the unpaired 3D pose samples}, $q_z$ from MADS~\cite{zhang2017martial} and CMU-MoCap~\cite{cmumocap} dataset keeping a clear domain gap with respect to the standard datasets chosen for benchmarking our performance.

\subsection{Evaluation on Human3.6M}\label{sec:4_1}\vspace{-1mm}
Inline with the prior arts~\cite{chen2019unsupervised,rhodin2018unsupervised}, we evaluate our 3D pose estimation performance in the standard protocol-II setting (\ie with scaling and rigid alignment). We experimented on 4 different variants of the proposed framework with increasing degrees of supervision levels. The base model in absence of any paired supervision is regarded as \textit{Ours(unsup)}. In presence of multi-view information (with camera extrinsics), we finetune the model by enforcing consistent canonical pose $p^{3D}$ and camera shift for multi-view image pairs, termed as 
\textit{Ours(multi-view-sup)}. Similarly, finetuning in presence of a direct supervision on the corresponding 2D pose GT is regarded as \textit{Ours(weakly-sup)}. Lastly, finetuning in presence of a direct 3D pose supervision on 10\% of the full training set is referred to as 
\textit{Ours(semi-sup)}. Table~\ref{tab:protocol2results} 
depicts our superior performance against the prior arts in their respective supervision levels.

\subsection{Evaluation on MPI-INF-3DHP}\vspace{-1mm}
With our framework, we also demonstrate a higher level of cross-dataset generalization, thereby 
minimizing the need for finetuning on novel unseen datasets. The carefully devised constraints at the intermediate pose representation are expected to restrict the model from producing implausible poses even when tested in unseen environments. To evaluate this, we directly pass samples of MPI-INF-3DHP~\cite{mehta2017monocular} (3DHP) test-set through \textit{Ours(weakly-sup)} model trained on YTube+H3.6M and refer it as unsupervised transfer, denoted as  
-3DHP in Table~\ref{tab:mpiinf3dhp}. Further, we finetune the \textit{Ours(weakly-sup)} 
on 3DHP dataset at three supervision levels, a) no supervision, b) full 2D pose supervision, and c) 10\% 3D pose supervision as reported in Table~\ref{tab:mpiinf3dhp}, at rows 10, 7, and 11 respectively. The reported metrics clearly highlight our superiority against the prior arts.

\begin{table}[t!]
	\caption{ 
	Comparison of 3D pose estimation results on Human3.6M. Comparable metrics of fully-supervised methods are included for reference. Our approach achieves state-of-the art performance while brought to the same supervision-level (divided by horizontal lines) of \textit{Full-2D} (row no. 4-8) or \textit{Mutli-view} (row no. 9-10). Moreover, \textit{Ours(semi-sup)} achieves comparable performance against the prior fully supervised approaches. \vspace{1mm}} 
	\centering
	\setlength\tabcolsep{8.3pt}
	\resizebox{0.48\textwidth}{!}{
	\begin{tabular}{ll|c|c}
	\hline
 		No. & Protocol-II & Supervision & 
 		Avg. MPJPE($\downarrow$) \\
		\hline\hline
		1. & Zhou \etal \cite{zhou2016sparse} & Full-3D & 
		106.7 \\
		2. & Chen~\etal~\cite{chen20173d} & Full-3D & 
		82.7 \\
		3. &Martinez \etal \cite{martinez2017simple} & Full-3D & 
		52.1 \\
		\hline

		4. & Wu \etal \cite{wu2016single} & Full-2D & 
		98.4 \\
		5. & Tung \etal \cite{tung2017adversarial} & Full-2D & 
		{97.2} \\
		
		6. & Chen \etal \cite{chen2019unsupervised} & Full-2D & 
		68.0 \\
		
		7. & Wandt \etal \cite{wandt2019repnet} & Full-2D & 
		65.1 \\
		
		 \rowcolor{gray!10}
		8. & \textit{Ours(weakly-sup)} & Full-2D & 
		\textbf{62.4} \\
		
        \hline
		9. & Rhodin \etal \cite{rhodin2018unsupervised} & {Multi-view} &
		98.2 \\ 
		
		\rowcolor{gray!10}
		10. & \textit{Ours(multi-view-sup)} & {Multi-view} & 
		\textbf{85.8} \\ \hline
		
		\rowcolor{gray!10}
		11. & \textit{Ours(unsup)} & \textbf{No sup.} & 
		{99.2} \\ 
        
        \rowcolor{gray!10}
		12. & \textit{Ours(semi-sup)} & 10\%-3D & 
		\textbf{50.8} \\
		\hline
	\end{tabular}}
	\label{tab:protocol2results}
\end{table} 

\begin{table}[t!]
	\footnotesize
	\caption{ 
	Ablation analysis, highlighting importance of various constraints and regularization in the proposed self-supervised 3D pose estimation framework. (Qualitative results in Fig.~\ref{fig:viewsyn}{\color{red}B}) \vspace{0.8mm}
	}
	\centering
	\setlength\tabcolsep{7.0pt}
	\resizebox{0.48\textwidth}{!}{
	\begin{tabular}{l|cc}
	\hline
 		\multirow{2}{*}{\makecell{Method\\ (unsup.)}}  & \multirow{2}{*}{\makecell{MPJPE($\downarrow$) on\\ Human3.6M}} & \multirow{2}{*}{\makecell{3DPCK($\uparrow$) on\\ MPI-3DHP}} \\\\
		\hline\hline
		\textit{Ours(unsup)} w/o ${T}_{fk}\circ T_c$ & 126.8 & 51.7 \\ 
		\textit{Ours(unsup)} w/o $q_z\sim\mathcal{D}_z$ & 178.9 & 40.3 \\
		\textit{Ours(unsup)} w/o $m_{sal}$ & 189.4 & 35.7 \\
		\rowcolor{gray!10}
		\textit{Ours(unsup)} & 99.2 & 77.4 \\
		
		\hline
	\end{tabular}}
	\vspace{-1mm}
	\label{tab:ablations}
\end{table} 

\begin{table}[t!]
	\footnotesize
	\caption{ 
	3D pose estimation on 3DHP. Here, 2nd column denotes whether the approach uses 3DHP samples (paired or unpaired) while training. And the 3rd column specifies the supervision level. 
	}
	\centering
	\setlength\tabcolsep{1.1pt}
	\resizebox{0.48\textwidth}{!}{
	\begin{tabular}{ll|cc|ccc}
	\hline
 		No.&Method & Trainset & 3DHP Sup. & PCK ($\uparrow$) & AUC ($\uparrow$) & MPJPE ($\downarrow$) \\
		\hline\hline
		1.& Mehta \etal \cite{mehta2017vnect} & +3DHP & Full-3D & 76.6 & 40.4 & 124.7 \\
		2.& Rogez \etal \cite{rogez2017lcr} & +3DHP & Full-3D & 59.6 & 27.6 & 158.4 \\		
		
		\hline
		3.& Zhou \etal \cite{zhou2017towards} & +3DHP & Full-2D & 69.2 & 32.5 & 137.1 \\
		
		4.& HMR \cite{kanazawa2018end} & +3DHP & Full-2D & 77.1 &	40.7 & 113.2 \\
		5.& Yang \etal \cite{yang20183d} & +3DHP & Full-2D & 69.0 & 32.0 & - \\
		6.& Chen \etal \cite{chen2019unsupervised} & +3DHP & Full-2D &   {71.7}   & {36.3}  & - \\
		
		\rowcolor{gray!10} 
		7.&\textit{Ours(weakly-sup)} & +3DHP & Full-2D & \textbf{84.6} & \textbf{60.8} & \textbf{93.9} \\
		\hline
		
		8.& Chen \etal \cite{chen2019unsupervised} & -3DHP & - &   {64.3}   & {31.6}  & - \\
		\rowcolor{gray!10}
		9.&\textit{Ours(weakly-sup)} & -3DHP & - & 82.1 & 56.3 & 103.8 \\ 
		\rowcolor{gray!10}
		10.&\textit{Ours(weakly-sup)} & +3DHP & \textbf{No sup.} & \textbf{83.2} & \textbf{58.7} & \textbf{97.6} \\ \hline
		\rowcolor{gray!10}
		11.&\textit{Ours(semi-sup)} & +3DHP & 10\%-3D & \underline{86.3} & \underline{62.0} & \underline{74.1} \\
		\hline
	\end{tabular}}
	\label{tab:mpiinf3dhp}
\end{table} 

\subsection{Ablation study}\label{sec:4_3}\vspace{-0.5mm}

\begin{figure*}[!tbhp]
\begin{center}
	\includegraphics[width=0.98\linewidth]{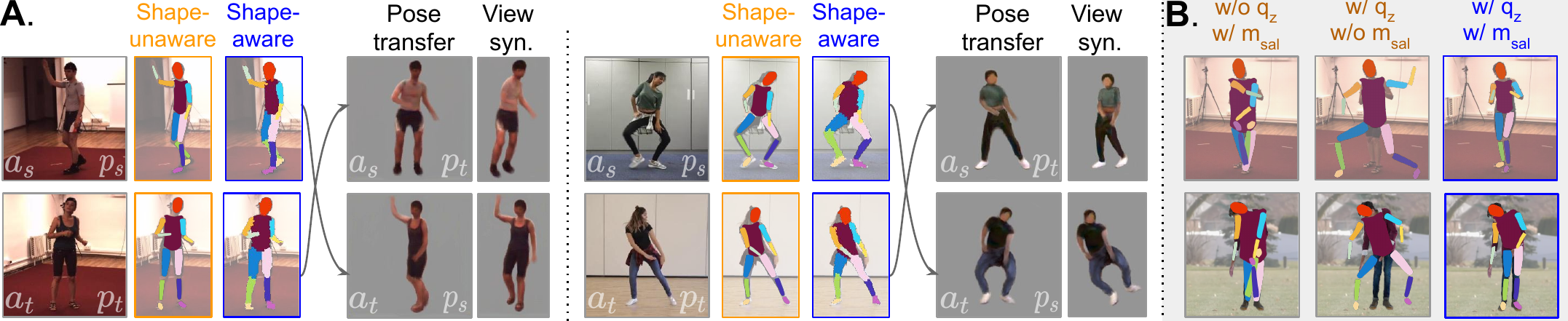}
	\vspace{-0.8mm}
	\caption{ 
	\textbf{A.} Novel image synthesis via latent manipulation of $a$, $p$ and $c$. It also shows the effect of independent non-rigid (\textit{pose-transfer}) and rigid (\textit{view-syn.}) variations as a result of explicit disentanglement. 
	Notice the corresponding shape-unaware (puppet deformation) and shape-aware part-segmentation results. \textbf{B.} Qualitative analysis of ablations showing importance of $q_z$ and $m_{sal}$ (refer Sec.~\ref{sec:4_3}).
	}
    \vspace{-4mm}
    \label{fig:viewsyn}  
\end{center}
\end{figure*}

We evaluate the effectiveness of the proposed local vector representation followed by the forward kinematic transformations, against a direct estimation of the 3D joints in camera coordinate system in the presence of appropriate bone length constraints~\cite{chen2019unsupervised}. As reported in Table~\ref{tab:ablations}, our disentanglement of camera from the view-invariant canonical pose shows a clear superiority as a result of using
the 3D pose articulation constraints in the most fundamental form. Besides this, we also perform ablations by removing $q_z$ or $m_{sal}$ from the unsupervised training pipeline. As shown in Fig.~\ref{fig:viewsyn}{\color{red}B}, without $q_z$ the model predicts implausible part arrangements even while maintaining a roughly consistent FG silhouette segmentation. However, without $m_{sal}$, the model renders a plausible pose on the BG area common between the image pairs, as a degenerate solution.  

\begin{figure}[!t]
\begin{center}
	\includegraphics[width=0.90\linewidth]{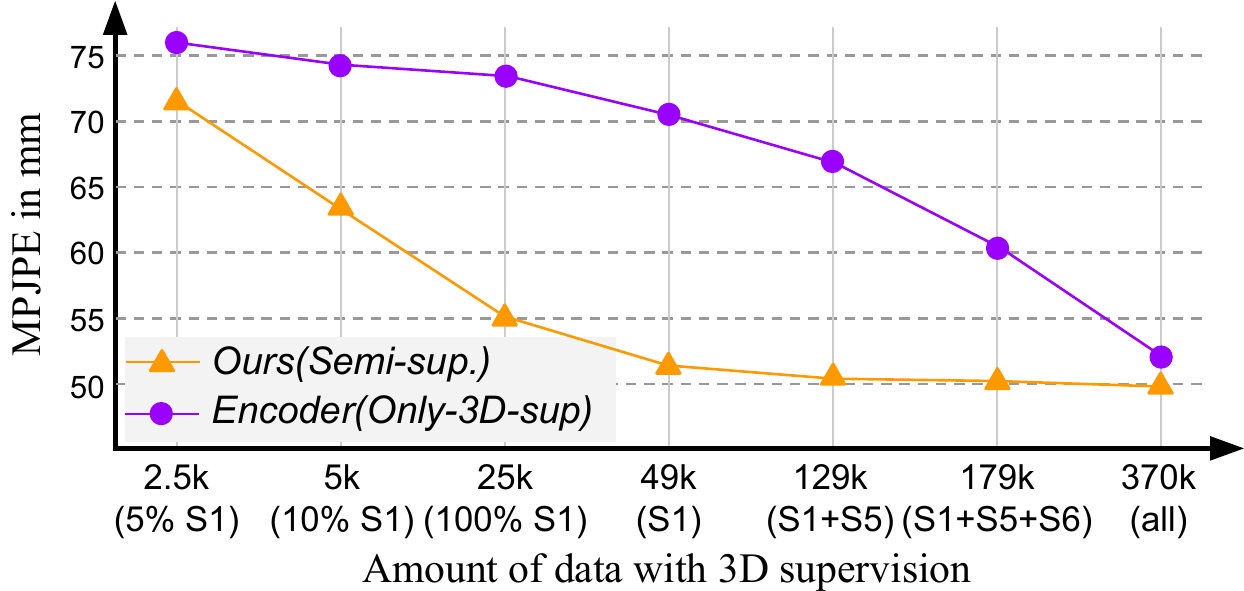}
	\caption{
	3D pose estimation on H3.6M as a function of the amount of training supervision. \textit{Ours(semi-sup)} shows faster transferability as compared to the fully supervised counterpart.
	}
    \vspace{-4mm}
    \label{fig:plot}  
\end{center}
\end{figure}

\begin{figure*}[!tbhp]
\begin{center}
    \includegraphics[width=0.98\linewidth]{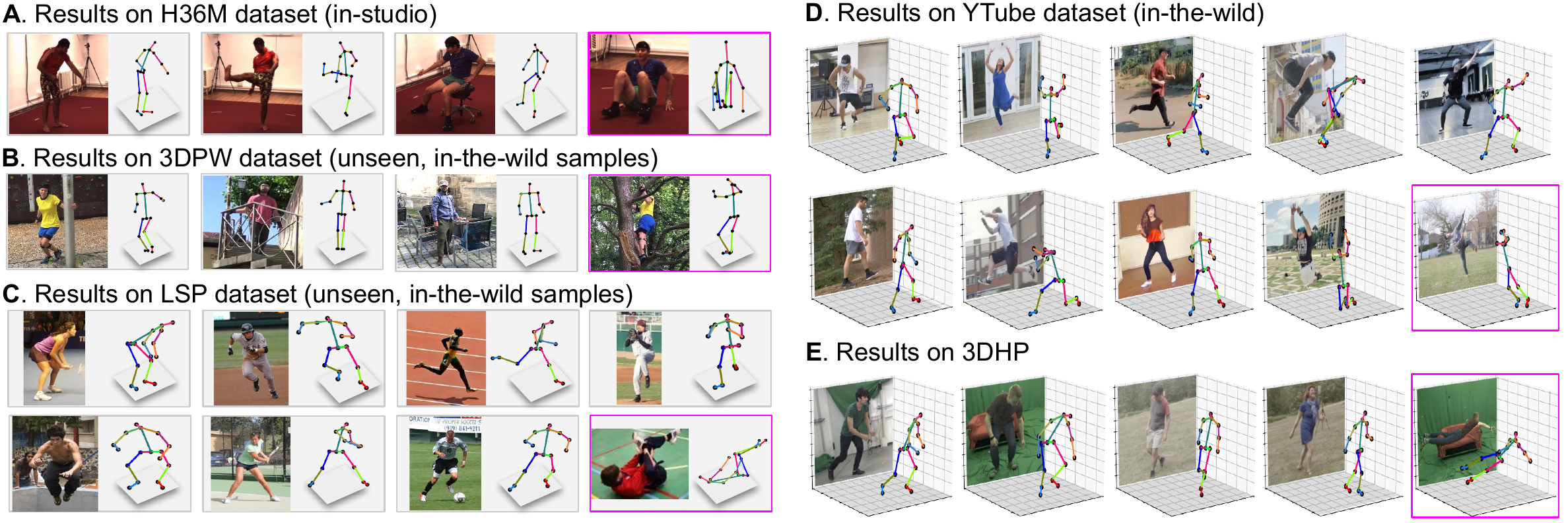}
	\vspace{-1mm}
	\caption{
	Qualitative results on 5 different datasets. Failure cases are highlighted in magenta which specifically occur in presence of multi-level inter-limb occlusion (see LSP failure case) and very rare, athletic poses (see YTube failure case). However, the model faithfully attends to single-level occlusions, enabled by the depth-aware part representation.
	}
    \vspace{-6mm}
    \label{fig:qualitative}  
\end{center}
\end{figure*}

As an ablation of the semi-supervised setting, (see Fig.~\ref{fig:plot}), 
we train the proposed framework on progressively increasing amount of 3D pose supervision alongside the unsupervised learning objective. Further, we perform the same for the Encoder network without the unsupervised objectives (thus, discarding the decoder networks) and term it as \textit{Encoder(Only-3D-sup)}. The plots in Fig.~\ref{fig:plot} clearly highlight our reduced dependency on the supervised data implying graceful and faster transferability.

\subsection{Evaluation of part segmentation}\vspace{-0.5mm}
For evaluation of part-segmentation, we standardize the ground-truth part conventions across both LSP~\cite{johnson2010clustered} and H3.6M datasets via SMPL model fitting~\cite{varol2017learning,LassnerClosing}. This convention roughly aligns with the part-puppet model used in Fig.~\ref{fig:main1}{\color{red}C}, thereby maintaining a consistent part to joint association. Note that, $w_{unc}$ is supposed to account for the ambiguity between the puppet-based shape-unaware segmentation against the image dependent shape-aware segmentation. In Fig.~\ref{fig:viewsyn}{\color{red}A}, we show the effectiveness of this design choice, where the shape-unaware segmentation is obtained at $y^p$ after depth-based part ordering, and the corresponding shape-aware segmentation is obtained at $\hat{y}$ output. Further, quantitative comparison of part segmentation 
is reported in Table~\ref{tab:part1}. 
We achieve comparable results against the prior arts, in absence of additional supervision.


\begin{table}[!t]
	\footnotesize
	\caption{Segmentation results on F1 metric ($\uparrow$) for LSP dataset.
	}
	\centering
	\setlength\tabcolsep{4.0pt}
	\resizebox{0.47\textwidth}{!}{
	\begin{tabular}{l|c|c|c}
	\hline
 		Method & Pose Sup. & FG vs BG & FG Parts \\
		\hline\hline
 		SMPLify~\cite{bogo2016keep} & Full-2D + \textit{SMPL-fitting} & 0.88  & 0.64    \\
  		\rowcolor{gray!00}
 		HMR~\cite{kanazawa2018end} & Full-2D + \textit{SMPL-fitting} & 0.86 & 0.59     \\
  		\hline \rowcolor{gray!10}
 		\textit{Ours(weakly-sup)} & Full-2D (no \textit{SMPL}) & 0.84 & 0.56   \\ 
 		\rowcolor{gray!10}
 		\textit{Ours(unsup)} & No sup. (no \textit{SMPL}) &0.78 &0.47   \\ 
		\hline
	\end{tabular}}
	\vspace{-2mm}
	\label{tab:part1}
\end{table} 



\subsection{Qualitative results}
To evaluate the effectiveness of the disentangled factors beyond the intended primary task of 3D pose estimation and part segmentation, we manipulate them to analyze their effect on the decoder synthesized output image. In \textit{pose-transfer}, pose obtained from an image is transferred to the appearance of another. However, in \textit{view-syn.}, we randomly vary the camera extrinsic values in $c$. The results shown in Fig.~\ref{fig:viewsyn}{\color{red}A} are obtained from \textit{Ours(unsup)} model, which is trained on the mixed YTube+H3.6 dataset.
This demonstrates the clear disentanglement of pose and appearance. 
Fig.~\ref{fig:qualitative} depicts qualitative results for the primary 3D pose estimation and part segmentation tasks using \textit{Ours(weakly-sup)} model 
as introduced in Sec.~\ref{sec:4_1}.
In Fig.~\ref{fig:qualitative}{\color{red}B}, we show results on the unseen LSP dataset, where the model has not seen this dataset even during self-supervised training. A consistent performance on such unseen dataset further establishes generalizability of the proposed framework. 

\section{Conclusion}\vspace{-1mm}

We proposed a self-supervised 3D pose estimation method that disentangles the inherent factors of variations via part guided human image synthesis. Our framework has two prominent traits. First, effective imposition of both human 3D pose articulation and joint-part association constraint via structural means. Second, usage of depth-aware part based representation to specifically attend to the FG human resulting in robustness to changing backgrounds. However, extending such a framework for multi-person or partially visible human scenarios remains an open challenge.

\vspace{1mm}\noindent
\textbf{Acknowledgements.} This project is supported
by a Indo-UK Joint Project (DST/INT/UK/P-179/2017), DST, Govt. of India and a Wipro PhD Fellowship (Jogendra).


{\small
\bibliographystyle{ieee_fullname}
\bibliography{egbib}
}

\end{document}